\documentclass{article}
\usepackage[table, dvipsnames]{xcolor}
\usepackage[utf8]{inputenc}
\usepackage{authblk}
\usepackage{setspace}
\usepackage[margin=1in]{geometry}
\usepackage{graphicx}
\graphicspath{ {./figures/} }
\usepackage{subcaption}
\usepackage{amsmath}
\usepackage{booktabs}
\usepackage{arydshln} 
\usepackage{multirow}
\usepackage{mathtools}  
\usepackage{xfrac}
\usepackage{wrapfig}
\usepackage{rotating}
\usepackage[absolute]{textpos}
\usepackage[style=nejm, 
citestyle=numeric-comp,
sorting=none]{biblatex}
\title{On the Status of Foundation Models for SAR Imagery}

\author{Nathan Inkawhich}

\affil{Air Force Research Laboratory}

\date{August 26, 2025}

\begin{document}
\begin{textblock}{15}(4.5,27)
Approved for Public Release; Distribution Unlimited. AFRL-2025-4494
\end{textblock}

\maketitle

\begin{abstract}
In this work we investigate the viability of foundational AI/ML models for Synthetic Aperture Radar (SAR) object recognition tasks.
We are inspired by the tremendous progress being made in the wider community, particularly in the natural image domain where frontier labs are training huge models on web-scale datasets with unprecedented computing budgets.
It has become clear that these models, often trained with Self-Supervised Learning (SSL), will transform how we develop AI/ML solutions for object recognition tasks -- they can be adapted downstream with very limited labeled data, they are more robust to many forms of distribution shift, and their features are highly transferable out-of-the-box.
For these reasons and more, we are motivated to apply this technology to the SAR domain.
In our experiments we first run tests with today's most powerful visual foundational models, including DINOv2 \cite{oquab2024dinov}, DINOv3 \cite{simeoni2025dinov3} and PE-Core \cite{pecore} and observe their shortcomings at extracting semantically-interesting discriminative SAR target features when used off-the-shelf.
We then show that Self-Supervised finetuning of publicly available SSL models with SAR data is a viable path forward by training several AFRL-DINOv2s and setting a new state-of-the-art for SAR foundation models, significantly outperforming today's best SAR-domain model SARATR-X.
Our experiments further analyze the performance trade-off of using different backbones with different downstream task-adaptation recipes, and we monitor each model's ability to overcome challenges within the downstream environments (e.g., extended operating conditions and low amounts of labeled data).
We hope this work will inform and inspire future SAR foundation model builders, because despite our positive results, we still have a long way to go.

\end{abstract}


\section{Introduction} \label{sec:introduction}





There has recently been significant effort in the AI/ML research community to develop foundational models.
A dominant trend is the scaling of both model size and dataset size, along with the use of Self-Supervised Learning (SSL) objectives.
In the image domain, we are now seeing training datasets on the order of several billion images and network architectures with billions of learnable parameters \cite{simeoni2025dinov3, pecore} (contrast this to yesterday's ResNet-50s \cite{HeZRS16} with 25M parameters trained on 1M ImageNet images). 
These pretrained foundation models are then used for many diverse downstream tasks via several task-adaptation workflows (e.g., as fixed feature extractors or with some finetuning).
For those in the Synthetic Aperture Radar (SAR) community, however, it remains an open question how applicable these public foundation models and associated techniques are to the SAR domain given that SAR data is not well-represented (if at all) in the web-based pretraining distributions.
Thus, our goals in this work are to (1) update the community on the efficacy of existing foundational AI/ML feature extractors for exploiting SAR imagery; (2) suggest useful SAR-domain evaluation techniques and task-adaptation practices for use with this emerging model type; and (3) present a promising direction for the design of future SAR-domain foundation models which involves self-supervised finetuning.


Since foundation models are a bit different than supervised ResNets of the past, we will now introduce some terminology that will be used throughout this paper.
First, it helps to think of visual foundation models trained with SSL as \textit{task-agnostic} feature extractors. 
They provide an embedding function for arbitrary images into a semantically-rich feature space where there are no explicit definitions of ``classes.''
This is notably different than \textit{task-specific} models trained with supervised learning which seek to bin all input images into one of $\mathcal{C}$ clusters, where $\mathcal{C}$ is the number of classes in the fully-labeled training dataset.
When we want to use a foundation model to solve a downstream task with $\mathcal{C}$ classes, we instantiate and train a \textit{task-head} to partition the model's feature space to accomplish the specific task.
A key advantage of well-trained foundation models is that often the feature extractor can remain fixed, and only the parameters of the task-head are learned. 
This significantly speeds up the task-adaptation process and often requires less labeled training data to learn the partition.
However, as you will see in this paper, it is sometimes beneficial to update the parameters of the feature extractor along with the task-head. 
Since there are many ways one can perform this task-adaptation, we will use the term \textit{task-adaptation recipe} to capture the steps and choices made during the process.

\begin{figure}[t]
    \centering
    \includegraphics[width=0.99\textwidth]{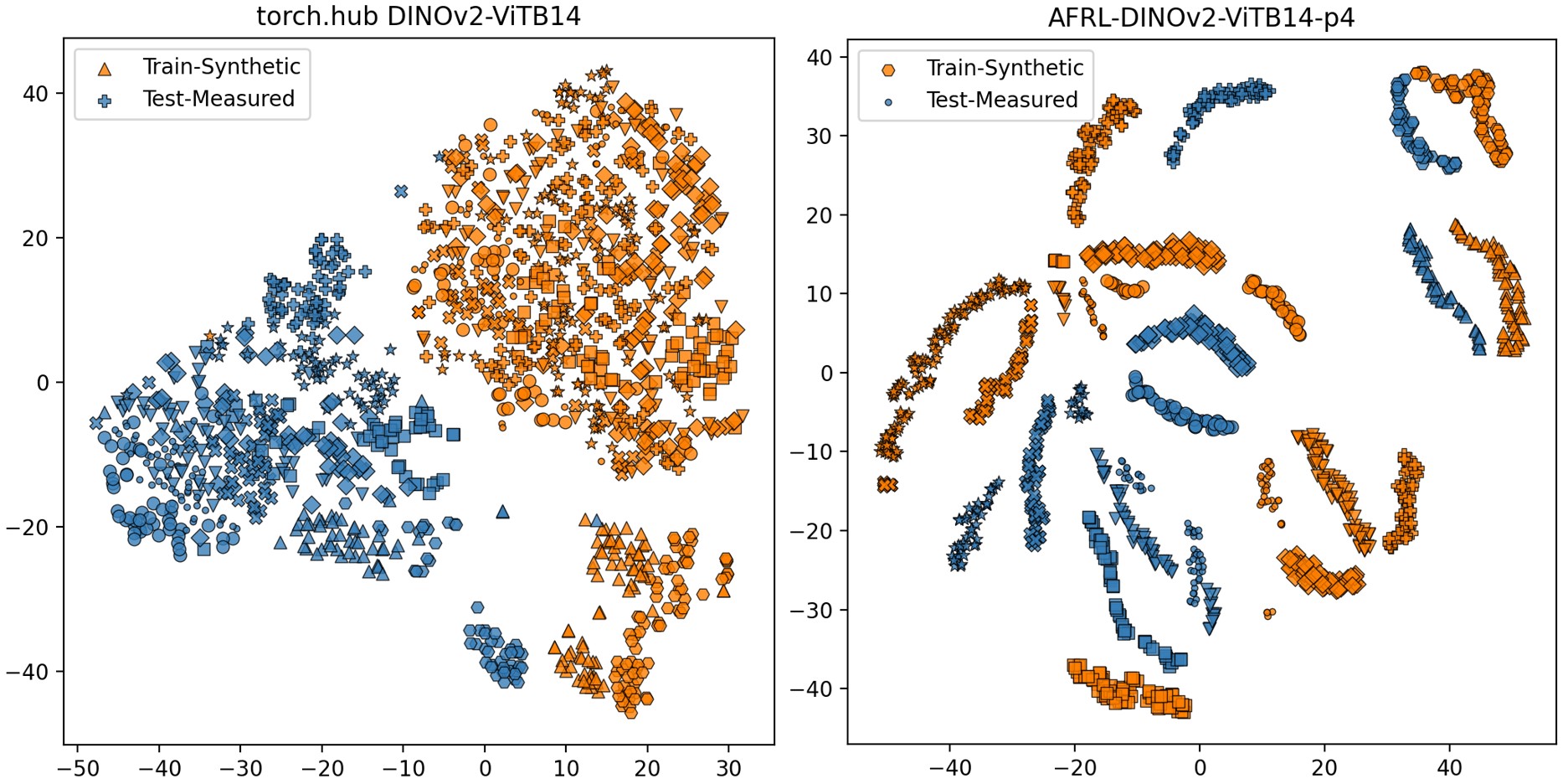}
    \caption{Feature space analysis between an off-the-shelf DINOv2 from \texttt{torch.hub} (left) and a custom SAR-finetuned DINOv2 (right) when embedding data from the SAMPLE (K=0)~\cite{sampledata} task (done with t-SNE).}
    \label{fig:ftspace-tsne}
\end{figure}

As a preview of what's to come, see Figure~\ref{fig:ftspace-tsne}.
In this experiment, we investigate the feature space of two DINOv2\footnote{DINOv2 is widely considered one of the most powerful SSL visual foundation model techniques available \cite{darcet2025capi, pecore}}~\cite{oquab2024dinov} foundation models when embedding data from the SAMPLE (K=0) task \cite{sampledata}, as visualized with t-SNE. 
Recall, SAMPLE involves the same 10 ground vehicle classes as MSTAR \cite{Ross1998MSTAR} and K=0 indicates the training data is fully synthetic while the test data is fully measured.
On the plots, the {\color{RedOrange}orange} markers are embeddings of the synthetic training data, the {\color{blue}blue} markers are embeddings of the measured test data, and the different marker shapes indicate different classes.
Finally, the embedding on the left subplot uses an off-the-shelf pretrained DINOv2 from \texttt{torch.hub} while the right subplot uses a custom SAR-finetuned DINOv2 we train in this work.
The difference is obvious.
The \texttt{torch.hub} DINOv2 (which has never been trained on a SAR image) produces an unusable feature space for this task which separates the data by synthetic vs.~measured instead of by class.
The SAR-finetuned DINOv2 produces a more nuanced and discriminative feature space which is separable by class and also co-locates many of the synthetic and measured representations of each class even though it has never been trained on a \textit{labeled} SAR image.
As will be discussed in more detail in the following sections, the implications of this observation are two-fold.
First, even the best publicly available foundation models from the natural image domain are not effective in the SAR domain off-the-shelf and require expensive task-adaptation recipes to be competitive downstream.
Second, SSL in the SAR domain is a promising approach to creating powerful SAR foundation models and these models can be used downstream with cheaper adaptation recipes.

Overall, our contributions in this work are as follows:
\begin{itemize}
    \item We provide a much needed update on the functionality of foundational AI/ML models for exploiting SAR data which includes today's most powerful visual models;
    \item We highlight the importance of specifying the downstream \textit{task-adaptation recipe} and show that there are several ways one can adapt a model depending on the suitability of the base model as well as compute and data limitations in the downstream environment;
    \item We make new comparisons between today's s.o.t.a. models from natural imagery (e.g., DINOv2~\cite{oquab2024dinov} and DINOv3~\cite{simeoni2025dinov3}) and several SAR-domain specific methods (e.g., SARATR-X~\cite{li2025saratr});
    \item Finally, we show a promising path forward for developing future SAR foundation models based on self-supervised finetuning DINOv2 with SAR data.
\end{itemize}

\section{Background \& Related Work} \label{sec:rw}

\textbf{Visual Foundation Models.}
By far, most of the development effort in visual foundation models has been in the natural imagery domain using web-based datasets. 
A key enabler of this model type, aside from the aggregation of humongous datasets, has been algorithmic advances in Self-Supervised Learning (SSL) -- a flavor of unsupervised learning that trains deep neural networks to capture semantically meaningful and discriminative features from unlabeled data.
Early versions of SSL models learned features by solving pretext tasks such as predicting rotations \cite{GidarisSK18}, vanilla contrastive learning \cite{simclr}, masked auto-encoding \cite{mae}, and forms of self-distillation \cite{byol, dinov1}.
Modern methods have built upon these and often hybridize learning concepts into composite training algorithms (e.g., DINOv2 \cite{oquab2024dinov, DarcetOMB24}, DINOv3 \cite{simeoni2025dinov3}, CAPI \cite{darcet2025capi}, PE-Core \cite{pecore}).
A full review of self-supervised learning is out of the scope of this work and we point readers to the SSL Cookbook \cite{sslcookbook}, helpful review papers \cite{10559458}, and the related work sections of today's best methods \cite{oquab2024dinov, simeoni2025dinov3, darcet2025capi, pecore}.

We also acknowledge some recent work towards SAR-specific foundation models. 
Note, the use of ``foundational'' here is quite loose given that the pretraining datasets are often $<<$1M samples and do not capture a significant diversity of targets and sensors, especially in comparison to the billion-scale web-based datasets used to pretrain frontier natural image models.
Inkawhich \cite{globalsar} trained a SimCLR-based \cite{simclr} model on SAR data and showed the benefits in few-shot learning and confuser rejection. The tests focus on MSTAR and a fixed feature extractor-based downstream usage recipe.
Li et al. \cite{li2024predicting} trained the SAR-JEPA model using a joint-embedding predictive architecture and showed the models flexibility at solving three different tasks involving ground vehicles, ships and aircraft. They consider both linear probing and full finetuning downstream task-adaptation recipes and both low-shot and full-shot data settings.
The same group later released the SARATR-X \cite{li2025saratr} model which is based on a Masked Image Modeling (MIM) SSL objective. 
Their pretraining recipe leverages finetuning of an ImageNet pretrained model, and further includes custom multi-scale gradient features and use of the HiViT \cite{zhang2023hivit} backbone. In \cite{li2025saratr} it is mentioned that SAR-JEPA \cite{li2024predicting} is a ``preliminary work'' and in the experiments highlight that ``SARATR-X outperform[s] our previous study SAR-JEPA.''
For this reason, we focus on comparison with the SARATR-X model in this work.
Also worth noting, this same group is responsible for the impressive ATRNet-STAR benchmark \cite{liu2025atrnet} which features extensive experiments and ultimately finds that the SARATR-X model performs the best across the 7 individual tasks.
Finally, we acknowledge the recently published SUMMIT SAR foundation model \cite{DU2025104624}.
SUMMIT also features a MIM SSL objective which is enhanced with several self-supervised auxiliary tasks.
In addition to the algorithmic improvements, SUMMIT also curates a large scale SAR pretraining dataset ($\sim$570k samples) which is a composite of many smaller public datasets including the ones used in this work (e.g., HRSID \cite{hrsid}, SSDD \cite{ssdd}, FUSAR \cite{fusar}, SRSDD \cite{srsdd}).
At the time of writing, the SUMMIT code and models are not publicly available. 
We leave it to future work to make comparisons with SUMMIT.

\textbf{DINOs outside of natural imagery.}
It's worth mentioning a few works that have used and/or adapted DINO-based models for tasks outside of the web-based natural imagery domain.
We draw inspiration from several papers that perform parameter-efficient adaptations of DINOv2s for the exploitation of medical images (e.g., X-Rays) \cite{10635887, beilei2024surgical}.
While the datasets and tasks (e.g., Lung Nodule Classification and Depth Estimation) are quite unique, SAR domain researchers can certainly learn from the finetuning configurations used.
Tolan et al.~\cite{tolan2024very} apply a DINOv2 to the problem of predicting tree canopy height maps from satellite images which advances our ability to understand carbon cycles and observe deforestation patterns.
Recently, a team in the 2024 CVPR PBVS MAVIC Challenge \cite{10678289} attempted to finetune a DINOv2 with Electro-optical (EO) and SAR data for the purpose of improving multi-modal classification performance \cite{abs-2412-12565}. Despite not ranking among the winning methods, we mention this because it's the first occurrence of a DINOv2 being adapted to the SAR domain that we are aware of.
Lastly, in the most recent frontier foundation model, DINOv3 \cite{simeoni2025dinov3}, there are several variants trained on EO satellite imagery from Maxar.
We will discuss their utility for SAR ATR in Section~\ref{sec:experiments}.

\textbf{Downstream usage recipes.}
Finally, we mention some common techniques that are used to adapt task-agnostic feature extractors to solve specific downstream tasks.
In the foundation model literature a common test to measure ``foundational-ness'' is to keep the feature extractor fixed and solve a variety of tasks with simple classifier head designs such as Nearest Neighbors and Linear Probes \cite{oquab2024dinov, simeoni2025dinov3, pecore}.
To measure peak s.o.t.a. performance models are usually finetuned for each downstream task separately.
An exciting research direction not always considered in visual foundation model papers is Parameter-Efficient Finetuning (PEFT). 
Methods such as Low Rank Adaptation (LoRA) \cite{hu2022lora} and SingLoRA \cite{singlora} introduce a small amount of extra learnable parameters that allow for adaptation of the feature extractors in an efficient way, making it possible to finetune very large vision transformers on modest hardware.
Finally, it is known that different SSL algorithms prefer different downstream recipes.
Specifically, MIM-based SSL algorithms, including Masked Auto-Encoders (MAEs) favor more expensive finetuning recipes and are known to under-perform the s.o.t.a. when used as fixed feature extractors \cite{darcet2025capi}.
DINOs and other joint-embedding models are known to work well as fixed feature extractors and when finetuned \cite{dinov1, oquab2024dinov, simeoni2025dinov3}.

\section{Finetuning DINOv2 for SAR} \label{sec:methods}

\begin{wrapfigure}{r}{.45\linewidth}
    \vspace{-4mm}
    \centering
    \includegraphics[width=0.99\linewidth]{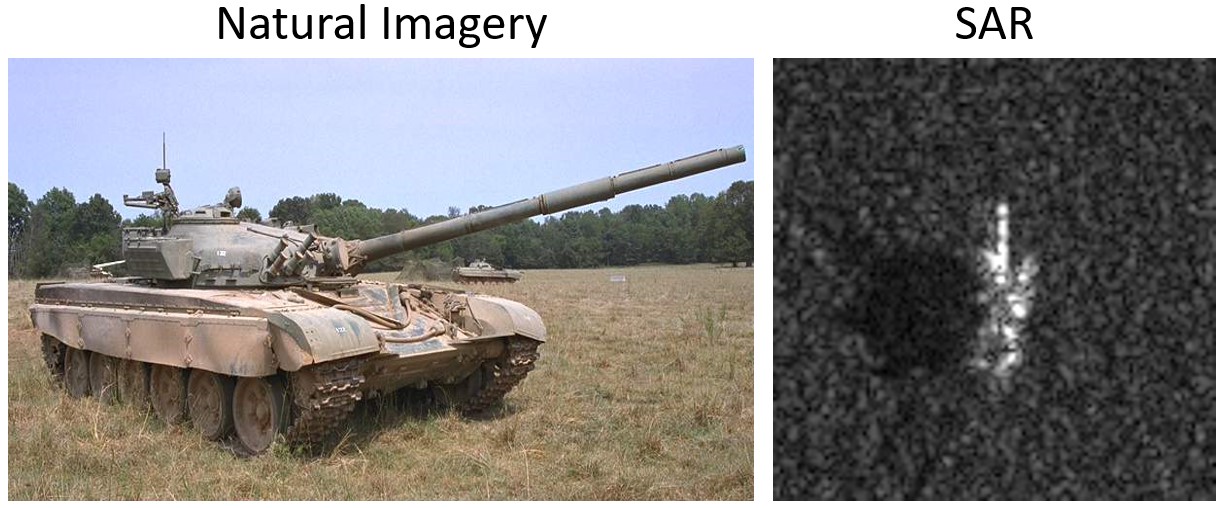}
    \caption{The same T-72 tank viewed in what we are calling the natural imagery domain (what you find on the web) vs as viewed in SAR.}
    \label{fig:eo-sar}
\end{wrapfigure}
The core development work in this project is the finetuning of a DINOv2 with SAR data.
Early on, we observed that off-the-shelf foundation models (e.g., \texttt{torch.hub} DINOv2s trained on internet data) were not well suited for use with SAR (ref.~Figure~\ref{fig:ftspace-tsne}). 
This isn't shocking as SAR images of tanks viewed from overhead are quite unique from cell phone pictures of tanks (Figure~\ref{fig:eo-sar}), as well as other miscellaneous objects that appear in the web-based pretraining sets such as German Shepherds, Tenchs and Taxi Cabs.
Frankly, there is no compelling reason to believe that models trained on web-based natural imagery will extract discriminative features of SAR targets.
Our approach to create a better foundation model for SAR is then to \textbf{finetune the \texttt{torch.hub} DINOv2s with a SAR dataset using the DINOv2 SSL objective}.

The most important aspect of this process is the curation of the SAR data for SSL pretraining.
A standard practice we follow is to aggregate separate datasets from across the internet into one large composite dataset \cite{oquab2024dinov,li2025saratr}.
We define our base pretraining set $\mathcal{D}_{pretrain}^{base}$ as the concatenation of SAR-Ships \cite{sarships}, OpenEarthMap \cite{openearthmap}, HRSID \cite{hrsid}, FUSAR \cite{fusar}, SSDD \cite{ssdd}, DualPolShips \cite{dualpolships}, SRSDD \cite{srsdd}, CVDome \cite{cvdome} and SARSIM \cite{sarsim}.
In total there are about 110,000 SAR images in $\mathcal{D}_{pretrain}^{base}$.
In some experiments we also include data from ATRNet-STAR \cite{liu2025atrnet} and SAMPLE \cite{sampledata}, the methodology for doing so is described next.

To test the impact of the inclusion/exclusion of a data distribution in the SSL $\mathcal{D}_{pretrain}$ we ultimately finetune four DINOv2s, all of which start from the \texttt{torch.hub dinov2\_vitb14\_reg} pretrained weights and use the same basic training configuration.
Our \texttt{AFRL-DINOv2-ViTB-p1} model is trained on $\mathcal{D}_{pretrain}^{base}$, the \texttt{AFRL-DINOv2-ViTB-p2} is trained on $\mathcal{D}_{pretrain}^{base}$+SAMPLE-syn, the \texttt{AFRL-DINOv2-ViTB-p3} is trained on $\mathcal{D}_{pretrain}^{base}$+ATRNet-SOC40-train; and the \texttt{AFRL-DINOv2-ViTB-p4} is trained on $\mathcal{D}_{pretrain}^{base}$+SAMPLE-syn+ATRNet-SOC40-train.
It is important to note that ATRNet-SOC40-train refers \textit{only} to the training set of the SOC-40 split of ATRNet-STAR~\cite{liu2025atrnet} dataset and SAMPLE-syn refers \textit{only} to the synthetic data included in the SAMPLE-public \cite{sampledata} dataset.
This will be important to keep in mind for evaluating the experimental results.

Since DINOv2 is a joint-embedding-style SSL algorithm, another important design detail is the data augmentations used to create diverse views.
We use an augmentation scheme inspired by \cite{globalsar}. 
Specifically, we use RandomResizedCrop, RandomRotation by 90$^\circ$, Random Horizontal and Vertical Flipping, ClipAndScale, PowScale, GaussianNoise, GaussianBlur, and the standard ImageNet normalization constants that the \texttt{torch.hub} DINOv2 was trained with.
We do not proclaim this to be the *best* augmentation recipe for SAR data, but we found it to be a reasonable baseline.
More research into clever SAR domain-specific augmentations for SSL pretraining is needed in future work.

Finally, we give some miscellaneous training details. 
We finetune for 100 epochs with a batch size of 512 and a base learning rate of 0.0004. 
As mentioned we use the ViTB14 backbone architecture with 4 register tokens \cite{DarcetOMB24}.
We set the global crop scale range to [0.6, 1.], the local crop scale range to [0.2, 0.6] and always use the default input pixel resolution of 224$\times$224.
To make the grayscale SAR images fit the 3-channel expected input, we simply replicate along the channel dimension.
In the future it is also interesting to investigate packing the channel dimension with various processed versions of the SAR image (e.g., wavelet transformed versions) as done in SARATR-X \cite{li2025saratr}.

\section{Experiments} \label{sec:experiments}

\subsection{Solving Downstream Tasks with Nearest Neighbor Classifiers} \label{sec:exp-knn}

As mentioned in Section \ref{sec:rw}, a popular way to empirically measure the quality of a visual foundation model is to keep the weights of the feature extractor fixed and train a very simple task-head to accomplish multiple diverse downstream tasks (often the head is a KNN or linear probe).
It then becomes interesting to look at average performance across tasks to get a sense of the model's expressiveness and flexibility (which are intuitively linked to \textit{foundational-ness}.)
The simple head design is meant to showcase the out-of-the-box performance of the foundation models at creating semantically meaningful groupings across heterogeneous tasks.
For the first experiment we opt to use a Nearest Neighbor classifier on top of many different fixed feature extractors to compare their performance on a benchmark of 10 downstream tasks.

\subsubsection{Setup}

\textbf{Downstream Task Datasets.~}
Three of the downstream tasks are from the 10-class MSTAR benchmarking family. We use the MSTAR-SOC and MSTAR-EOC splits from \cite{globalsar}, where SOC involves a 2$^\circ$ elevation shift between train/test and EOC involves a 13$^\circ$ shift (SOC = Standard Operating Conditions and EOC = Extended Operating Conditions).
We also test on the SAMPLE~(K=0) task \cite{sampledata} which uses 100\% synthetically generated data for training and 100\% measured data for testing.
The other seven downstream tasks are from the recently released ATRNet-STAR benchmark \cite{liu2025atrnet}. 
This well-structured dataset includes 40 classes of ground vehicles including cars, busses, trucks and special vehicles such as tractors collected under many different scenarios (e.g., different scenes, at different geometries, and with different frequency bands and polarizations).
We use the exact Ground Plane splits from the official release, named SOC-40, SOC-50, EOC-Scene, EOC-Depression, EOC-Azimuth, EOC-Band and EOC-Polarization.
We encourage the reader to check out \cite{liu2025atrnet} for full details but mention that the names give away key information. 
In short, SOC-40 is a 40 class task where the train/test data are sampled i.i.d.~from the same set of conditions and SOC-50 is similar but with 50 classes.
All of the EOCs are 40 classes. In EOC-Scene the train samples have simple background clutter while the test samples have complex clutter; EOC-Band uses X-band data for training and Ku-band for test; EOC-Azimuth restricts train views to [0$^\circ$, 60$^\circ$) and test to [60$^\circ$, 360$^\circ$); EOC-Depression restricts train depressions to 15$^\circ$ and test to [30$^\circ$, 45$^\circ$, 60$^\circ$]; and EOC-Polarization trains with HH and tests on HV, VH and VV.

\textbf{Feature Extractors.~}
We consider 16 different feature extractors which come from three basic groupings.
The first group contains models from the wider AI/ML community that are trained on web-data (not SAR).
As a control, we include an ImageNet-trained ResNet50 from torchvision which is the only feature extractor trained with supervised learning.
The rest of the models in this group are widely considered the most powerful visual foundation models in the world today and include DINOv2s \cite{oquab2024dinov}, DINOv3s \cite{simeoni2025dinov3}, and PE-Cores \cite{pecore} that were obtained via \texttt{torch.hub} and \texttt{huggingface}.
Since scaling is so important to the performance of these models (usually bigger is better) we grab a sampling of model sizes, from ViT-Base with 86M parameters up to ViT-7b with nearly 7B parameters.
The second group of models come from the SAR community and are early versions of SAR foundation models.
We consider two SARATR-X \cite{li2025saratr} checkpoints, one released with the paper and the other released with ATRBench \cite{liu2025atrnet}.
At the time of writing, SARATR-X is the newest and most performant foundational SAR model we are aware of.
We also use the SAR-SimCLR from \cite{globalsar}.
Lastly, the third group of models contains the four SAR-finetuned AFRL-DINOv2s mentioned in Sec.~\ref{sec:methods}.

\subsubsection{Results}
Table~\ref{tab:main-knn} contains the main results for this experiment.
Each row corresponds to one of the aforementioned pretrained feature extractors, with \# parameters noted.
Each column maps to one of the 10 downstream tasks. 
Since some of the tasks have an unbalanced number of test images per class we report the balanced accuracy statistic.
The rightmost column shows the average accuracy across the 10 tasks.
Finally, \colorbox{orange!20}{orange} shading in a box indicates that the feature extractor has seen \textit{unlabeled} samples from the task distribution during SSL pretraining.
\footnote{A discussion on the implications of this assumption is in Appendix A.}

The first observation we make is that models that have been updated on at least some SAR data (groups 2 and 3) are more performant than models trained only on natural imagery, which verifies a rather obvious intuition but is a good sanity check.
Among the group of public web-based models we do see some benefits to scaling model size -- the DINOv3-ViT7b-Web is generally the highest performing -- but for the orders of magnitude more parameters we only get a few percentage points boost on average.
Another interesting comparison among the public models is between DINOv3-ViTL-Web and DINOv3-ViTL-Sat. 
The former was trained on a web-based dataset with $\sim$1.7B samples while the latter was trained on a satellite image dataset with $\sim$500M samples from Maxar.
Intuitively, we might expect that the -Sat model may do better because the pretraining domain (i.e., remote sensing) is closer to our SAR domain, however, this is not the case.
We believe this is because the satellite data encourages the model to learn more ``scene'' relevant features whereas the web data encourages the learning of ``object'' relevant features and our downstream benchmark tasks are mainly object-centric recognition tasks.

\begin{table}[t] 
\caption{Accuracy results for 10 different downstream tasks using a Nearest Neighbor task-head.}
\centering
\label{tab:main-knn}
\resizebox{1.\textwidth}{!}{
\begin{tabular}{lcccccccccccc}
\toprule
                           &                                                                           & \multicolumn{3}{c}{MSTAR Family} & \multicolumn{7}{c}{ATRNet-STAR Benchmark}                            &       \\ \cmidrule(lr){3-5} \cmidrule(lr){6-12}
Pretrained Ft Extractor    &  \rotatebox{90}{\# params (M)}                                                                         & \rotatebox{90}{MSTAR-SOC}     & \rotatebox{90}{MSTAR-EOC}    & \rotatebox{90}{SAMPLE-K0}    & \rotatebox{90}{SOC-40} & \rotatebox{90}{SOC-50} & \rotatebox{90}{EOC-az} & \rotatebox{90}{EOC-scene} & \rotatebox{90}{EOC-pol} & \rotatebox{90}{EOC-depr} & \rotatebox{90}{EOC-band} & \cellcolor{gray!20}AVG   \\ \midrule
Torchvision RN50           & 24                                                                        & 48.6          & 57.7         & 20.2         & 35.7   & 27.1   & 12.2   & 6.6       & 22.8    & 10.3     & 27.1     & \cellcolor{gray!20}26.8 \\
torch.hub DINOv2-ViTB      & 86                                                                        & 64.5          & 52.4         & 20.6         & 38.7   & 32     & 13.5   & 6.2       & 25.6    & 11.5     & 30.5     & \cellcolor{gray!20}29.5 \\
torch.hub DINOv2-ViTL      & 304                                                                       & 64.2          & 53.4         & 23.9         & 35.9   & 30.6   & 13     & 5.7       & 23.6    & 11.2     & 28.4     & \cellcolor{gray!20}28.9 \\
torch.hub DINOv2-ViTG      & 1,130                                                                      & 64.8          & 50.4         & 24.7         & 41.3   & 34.1   & 14.6   & 6.2       & 26.6    & 12.4     & 32.6     & \cellcolor{gray!20}30.7 \\
torch.hub DINOv3-ViTL-Web  & 304                                                                       & 69.2          & 54.4         & 25.3         & 38.3   & 32.5   & 14     & 5.8       & 25.8    & 12       & 31.5     & \cellcolor{gray!20}30.8 \\
torch.hub DINOv3-ViTL-Sat  & 304                                                                       & 51            & 47.3         & 23.7         & 36     & 28.1   & 10.6   & 5.7       & 24.2    & 8.2      & 21.5     & \cellcolor{gray!20}25.6 \\
torch.hub DINOv3-ViT7b-Web & 6,716                                                                      & 72.9          & 56.1         & 33.4         & 44     & 36.8   & 15.8   & 6.3       & 29.8    & 13.3     & 35.6     & \cellcolor{gray!20}34.4  \\
hf/timm/PE-Core-L14-336    & 316                                                                       & 65.6          & 58.6         & 23.2         & 40.2   & 32.8   & 14.4   & 7.4       & 27      & 12.5     & 30.5     & \cellcolor{gray!20}31.2 \\
hf/timm/PE-Core-G14-448    & 1,880                                                                      & 65.4          & 56.6         & 29.6         & 38.7   & 31     & 13.6   & 6.3       & 25.2    & 11.5     & 29.5     & \cellcolor{gray!20}30.7 \\ \midrule
SARATR-X-notest            & 66                                                                        & {\cellcolor{orange!20}98.7}          & {\cellcolor{orange!20}77.3}         & {\cellcolor{orange!20}48.3}         & 56.6   & 43.4   & 11     & 5.4       & 37.5    & 11.9     & 42.7     & \cellcolor{gray!20}43.2 \\
SARATR-X-ckpt800           & 66                                                                        & {\cellcolor{orange!20}\textbf{99.3}}          & {\cellcolor{orange!20}82.8}         & {\cellcolor{orange!20}54.5}         & 57.2   & 43.8   & 11     & 5.5       & 38.9    & 12.7     & 40.7     & \cellcolor{gray!20}44.6 \\
SAR-SimCLR                 & 11                                                                        & 98.4          & 79.4         & 65           & 84.7   & 66.8   & 21.4   & 16.9      & 65.9    & 26.3     & 78.5     & \cellcolor{gray!20}60.3 \\ \midrule
AFRL-DINOv2-ViTB-p1        & 86                                                                        & 96.5          & 86.4         & 59.5         & 84.1   & 67.7   & 31.1   & 20.8      & 66.3    & 26.8     & 77.1     & \cellcolor{gray!20}61.6 \\
AFRL-DINOv2-ViTB-p2        & 86                                                                        & 97.4          & 86.1         & {\cellcolor{orange!20}57.9}         & 84.7   & 68.2   & 31.5   & 21.3      & 66.4    & 26.7     & 77.4     & \cellcolor{gray!20}61.7 \\
AFRL-DINOv2-ViTB-p3        & 86                                                                        & 98.9          & 90.7         & 72.9         & \cellcolor{orange!20}\textbf{96.7}   & \cellcolor{orange!20}87.2   & \cellcolor{orange!20}\textbf{59.7}   & \cellcolor{orange!20}\textbf{45.2}      & \cellcolor{orange!20}94      & \cellcolor{orange!20}\textbf{57.8}     & \cellcolor{orange!20}\textbf{93.4}     & \cellcolor{gray!20}\textbf{79.6} \\
AFRL-DINOv2-ViTB-p4        & 86                                                                        & 99            & \textbf{92.4}         & \cellcolor{orange!20}\textbf{73.9}         & \cellcolor{orange!20}\textbf{96.7}   & \cellcolor{orange!20}\textbf{87.3}   & \cellcolor{orange!20}59.1   & \cellcolor{orange!20}43.7      & \cellcolor{orange!20}\textbf{94.1}    & \cellcolor{orange!20}57.6     & \cellcolor{orange!20}93       & \cellcolor{gray!20}\textbf{79.6} \\ \bottomrule
\end{tabular}
}
\end{table}

Among the existing SAR community models the SAR-SimCLR from \cite{globalsar} is quite impressive, especially for its size. 
The SARATR-X's do not perform particularly well for how much effort was put into the pretraining. 
As previously commented on, we believe this is because of the well-documented behavior that models pretrained with MIM SSL objectives do not perform well as fixed feature extractors \cite{darcet2025capi}. 

Our AFRL-DINOv2s are generally the best models in this experiment. 
Unsurprisingly, as we include more data in $\mathcal{D}_{pretrain}$ the performance improves (we're up to almost 80\% average accuracy with -p3 and -p4).
We also observe some more nuanced behaviors.
First, note the performance change between -p1 and -p2 on the SAMPLE~(K=0) task -- accuracy actually \textit{decreases} slightly despite adding SAMPLE data into the pretraining set. 
We do not think it's productive to read into this much for three reasons: (1) the amount of SAMPLE data added to $\mathcal{D}_{pretrain}$ is only $\sim$1.3k samples, which is a very small portion of the overall set and we would not expect huge performance implications either way; (2) the SAMPLE data added to $\mathcal{D}_{pretrain}$ is fully-synthetic yet the SAMPLE~(K=0) task tests on measured data, so there is little expectation that the model would have learned to overcome such a distribution shift during SSL training; and (3) SSL training is a highly stochastic and there is sure to be variance (the performance $\Delta$ we are talking about is only 1.6\%).
Interestingly, if we look at the SAMPLE~(K=0) performance between -p2 and -p3 we see an impressive 15\% boost even though the -p3 model was not pretrained on any SAMPLE data.
This highlights how important it is at this stage of SAR foundation model development to continue to add large and diverse datasets to $\mathcal{D}_{pretrain}$ (ATRNet-SOC40-train added 68k samples).
Finally, we want to draw attention to the impressive robustness of the AFRL-DINOv2s, particularly the -p3 and -p4, on the ATRNet EOC settings despite the backbone having never been trained on a labeled image from that distribution.
This suggests the feature space of the SSL model is organized in a semantically useful way.

\subsection{``Trying Harder'' with Different Task-Adaptation Recipes} \label{sec:exp-recipes}

We acknowledge that nearest neighbor classifiers on fixed feature extractors may not be how serious SAR practitioners use these models to chase s.o.t.a. performance.
Instead, depending on the hardware setup, timelines, and data availability, developers may ``try-harder'' to adapt the pretrained models to their downstream tasks by introducing new parameters to be trained and/or actually finetuning parts of the backbone feature extractor.
We described this in the Introduction as using a more complex downstream \textit{task-adaptation recipe}.

In this section we consider five different adaptation recipes of varying complexity.
We consider the same suite of benchmark tasks and pick four pretrained models to use as foundational backbones.
The first three task-adaptation methods do \textit{not} attempt to modify any aspects of the feature extractor and instead only consider different ``task-head'' designs.
The first method is the \textbf{Nearest Neighbor} classifier from the previous experiment and is considered the baseline here.
The second method is the \textbf{Linear Probe} where we instantiate a single fully-connected layer which inputs the raw feature vectors and outputs predictions over the classes. The weights of the probe (only) are trained with stochastic gradient descent using a supervised learning objective.
The third method is the \textbf{Multi-layer Probe} which is similar to the Linear Probe except we use two fully-connected layers which allow learning of non-linear decision boundaries.
The fourth method trains \textbf{LoRA} adapters on all of the attention blocks in the backbone and uses the Multi-layer Probe head to produce predictions over the class sets. 
We use a LoRA rank of 3 which introduces a very limited amount of extra trainable parameters (code inspired by \cite{2024dinov2_lora_seg}).
The fifth method does \textbf{Full-Finetuning} of all backbone weights in conjunction with a Multi-layer Probe head.
This is by far the most complex in terms of compute, update time, and number of learnable parameters but is a common strategy for task adaptation today.
For all methods except Nearest Neighbor we use the same task-specific training setup: 100 epochs, ADAM optimizer with learning rate of 0.001, Cosine learning rate decay schedule, and supervised cross-entropy loss with label smoothing $\alpha$=0.1.

Table~\ref{tab:try-harder} contains the results of this experiment.
Note the third column shows $\Delta$ params (M) which counts the number of trainable parameters involved in the task-adaptation as measured in millions. 
For each backbone we \textbf{bold} the best performing task adaptation strategy and we use a \colorbox{green!20}{green} highlight to indicate the best results for each task over all configs.

Across all backbones we see a common trend that the highest performing recipes tend to be either LoRA or Full-Finetuning.
This isn't particularly surprising because there is a relatively small amount of data used for SAR SSL pretraining ($<<$ 1 Million) so the quality of the off-the-shelf features may be limited.
We expect that in the future as SAR $\mathcal{D}_{pretrain}$ sets grow to include Millions+ of diverse samples the models will be much more competitive for use as fixed feature extractors when chasing s.o.t.a.

Among the methods that do not attempt to modify the backbone, Multi-layer Probe is consistently the highest performing.
We see that this probe boosts the performance of \texttt{torch.hub} DINOv2 and SARATR-X more than the AFRL-DINOv2s (when compared to the Nearest Neighbor baselines).
We intuit that these models have learned a much weaker representation of SAR data and thus require more extensive update techniques to be competitive.
Another global trend across backbones is that LoRA finetuning gives all models a notable performance bump over the Multi-layer Probe.
This boost is most pronounced on the weaker \texttt{torch.hub} DINOv2 and SARATR-X models but is still clear on the AFRL-DINOv2s.
Perhaps the most impressive part of LoRA finetuning is that for the large performance boost we add an almost negligible amount of trainable parameters compared to the Multi-layer Probe and especially compared to Full-Finetuning.
That is a key takeaway of this experiment -- LoRA finetuning for downstream task-adaptation gives a great tradeoff between adaptation effort and performance gain and should be explored as an option if the downstream hardware allows.

\begin{table}[]
\caption{Accuracy results when using more complex task adaptation recipes.}
\label{tab:try-harder}
\centering
\resizebox{1.\textwidth}{!}{
\begin{tabular}{llcccccccccccc}
\toprule
                                            &                                   &                         & \multicolumn{3}{c}{MSTAR Family}  & \multicolumn{7}{c}{ATRNet-STAR Benchmark}                            &       \\ \cmidrule(lr){4-6} \cmidrule(lr){7-13}
\multicolumn{1}{c}{Pretrained Ft Extractor} & \multicolumn{1}{c}{Task Strategy}  &  \rotatebox{90}{$\Delta$ params (M)}                                                                         & \rotatebox{90}{MSTAR-SOC}     & \rotatebox{90}{MSTAR-EOC}    & \rotatebox{90}{SAMPLE-K0}    & \rotatebox{90}{SOC-40} & \rotatebox{90}{SOC-50} & \rotatebox{90}{EOC-az} & \rotatebox{90}{EOC-scene} & \rotatebox{90}{EOC-pol} & \rotatebox{90}{EOC-depr} & \rotatebox{90}{EOC-band} & \cellcolor{gray!20}AVG   \\ \midrule
\multirow{5}{*}{torch.hub DINOv2-ViTB}      & Nearest Neighbor                  & 0                       & 64.5      & 52.4      & 20.6      & 38.7   & 32     & 13.5   & 6.2       & 25.6    & 11.5     & 30.5     & \cellcolor{gray!20}29.5 \\
                                            & Linear Probe                      & 0.03                    & 72.9      & 65.4      & 31.8      & 37.4   & 42.4   & 19     & 8.8       & 29.4    & 18.6     & 40.6     & \cellcolor{gray!20}36.6 \\
                                            & Multi-layer Probe                 & 1.24                    & 91.6      & 75.8      & 37.5      & 63.7   & 53.5   & 21.9   & 8.6       & 44.7    & 21.4     & 56.3     & \cellcolor{gray!20}47.5  \\
                                            & LoRA Finetune                & 1.35                    & \textbf{99.3}      & 84.8      & \textbf{58.8}      & 98.6   & 89.3   & \textbf{34.8}   & 23        & 87.7    & \textbf{45.6}     & 95.2     & \cellcolor{gray!20}\textbf{71.7} \\
                                            & Full-Finetune                     & 87                      & 97.7      & \textbf{85.2}      & 56.5      & \textbf{99}     & \textbf{89.4}   & 24.2   & \textbf{24.3}      & \textbf{89.1}    & 43.8     & \textbf{95.9}     & \cellcolor{gray!20}70.5 \\ \midrule
                                            
\multirow{5}{*}{SARATR-X-notest}            & Nearest Neighbor                  & 0                       & 98.7      & 77.3      & 48.3      & 56.6   & 43.4   & 11     & 5.4       & 37.5    & 11.9     & 42.7     & \cellcolor{gray!20}43.2 \\
                                            & Linear Probe                      & 0.02                    & 98.3      & 72.8      & 67.5      & 37.4   & 44.3   & 13.5   & 8.6       & 28.7    & 17       & 39.6     & \cellcolor{gray!20}42.7 \\
                                            & Multi-layer Probe                 & 0.56                    & 99.4      & 91.6      & 81.1      & 74.8   & 62.3   & 14.8   & 8.2       & 50.4    & 20.8     & 64.8     & \cellcolor{gray!20}56.8 \\
                                            & LoRA Finetune                & 0.69                    & \textbf{99.6}      & 93.7      & 84.9      & 97.7   & 85.2   & 24.9   & 14.6      & 81.6    & 38.8     & 92.5     & \cellcolor{gray!20}71.3 \\
                                            & Full-Finetune                     & 66                      & \textbf{99.6}      & \textbf{94.1}      & \cellcolor{green!20}\textbf{89}        & \textbf{99.2}   & \textbf{90.6}   & \textbf{28.7}   & \textbf{21.6}      & \textbf{87.9}    & \textbf{45.1}     & \textbf{95.9}     & \cellcolor{gray!20}\textbf{75.1} \\ \midrule
                                            
\multirow{5}{*}{AFRL-DINOv2-ViTB-p1}        & Nearest Neighbor                  & 0                       & 96.5      & 86.4      & 59.5      & 84.1   & 67.7   & 31.1   & 20.8      & 66.3    & 26.8     & 77.1     & \cellcolor{gray!20}61.6 \\ 
                                            & Linear Probe                      & 0.03                    & 95.2      & 90.8      & 46        & 63.1   & 63.3   & 29.5   & 18        & 50.4    & 27       & 64.6     & \cellcolor{gray!20}54.7 \\
                                            & Multi-layer Probe                 & 1.24                    & 99.2      & 93.9      & 54.5      & 87.1   & 73.5   & 35.1   & 19.8      & 65.2    & 30.2     & 80.2     & \cellcolor{gray!20}63.8 \\
                                            & LoRA Finetune                & 1.35                    & \textbf{99.9}      & \textbf{95.7}      & 67.9      & 98.2   & 87.4   & \textbf{36.6}   & \textbf{27.1}      & 84.5    & 44.1     & 94       & \cellcolor{gray!20}73.5 \\
                                            & Full-Finetune                     & 87                      & 99.4      & 94        & \textbf{72.4}      & \textbf{99.2}   & \textbf{91.3}   & 29.1   & 25.2      & \textbf{90.5}    & \textbf{46.7}     & \textbf{97}       & \cellcolor{gray!20}\textbf{74.4} \\ \midrule
                                            
\multirow{5}{*}{AFRL-DINOv2-ViTB-p3}        & Nearest Neighbor                  & 0                       & 98.9      & 90.7      & 72.9      & 96.7   & 87.2   & 59.7   & 45.2      & 94      & 57.8     & 93.4     & \cellcolor{gray!20}79.6 \\
                                            & Linear Probe                      & 0.03                    & 98.6      & 93.3      & 74.1      & 90.1   & 87.2   & 57     & 30.5      & 84.1    & 49.2     & 89.8     & \cellcolor{gray!20}75.3 \\
                                            & Multi-layer Probe                 & 1.24                    & 99.8      & \cellcolor{green!20}\textbf{97.3}      & 71.6      & 98.3   & 92     & 63.6   & 40.1      & 92.8    & 57.7     & 95.3     & \cellcolor{gray!20}80.8 \\
                                            & LoRA Finetune                & 1.35                    & \cellcolor{green!20}\textbf{100}       & 96.5      & \textbf{79.3}      & 99.4   & \cellcolor{green!20}\textbf{94.9}   & \cellcolor{green!20}\textbf{66}     & \cellcolor{green!20}\textbf{47.6}      & \cellcolor{green!20}\textbf{95.5}    & \cellcolor{green!20}\textbf{63.8}     & 97.6     & \cellcolor{green!20}\textbf{84.0} \\
                                            & Full-Finetune                     & 87                      & 99.4      & 94.6      & 78.5      & \cellcolor{green!20}\textbf{99.5}   & 94.7   & 43.7   & 33.9      & 94.7    & 54.3     & \cellcolor{green!20}\textbf{97.7}     & \cellcolor{gray!20}79.1 \\ \bottomrule
\end{tabular}
}
\end{table}

For interested readers, in Appendix B we provide a more in-depth comparison of our results to the ones presented in \cite{liu2025atrnet}. In short, we confirm that the SARATR-X results presented here are faithful and that our AFRL-DINOv2s are infact more effective on all of the the ATRNet-STAR benchmark tasks across the board making them the new state-of-the-art.

\subsection{Low-shot Experiments} \label{sec:exp-lowshot}

\begin{figure}[t]
    \centering
    \includegraphics[width=0.99\textwidth]{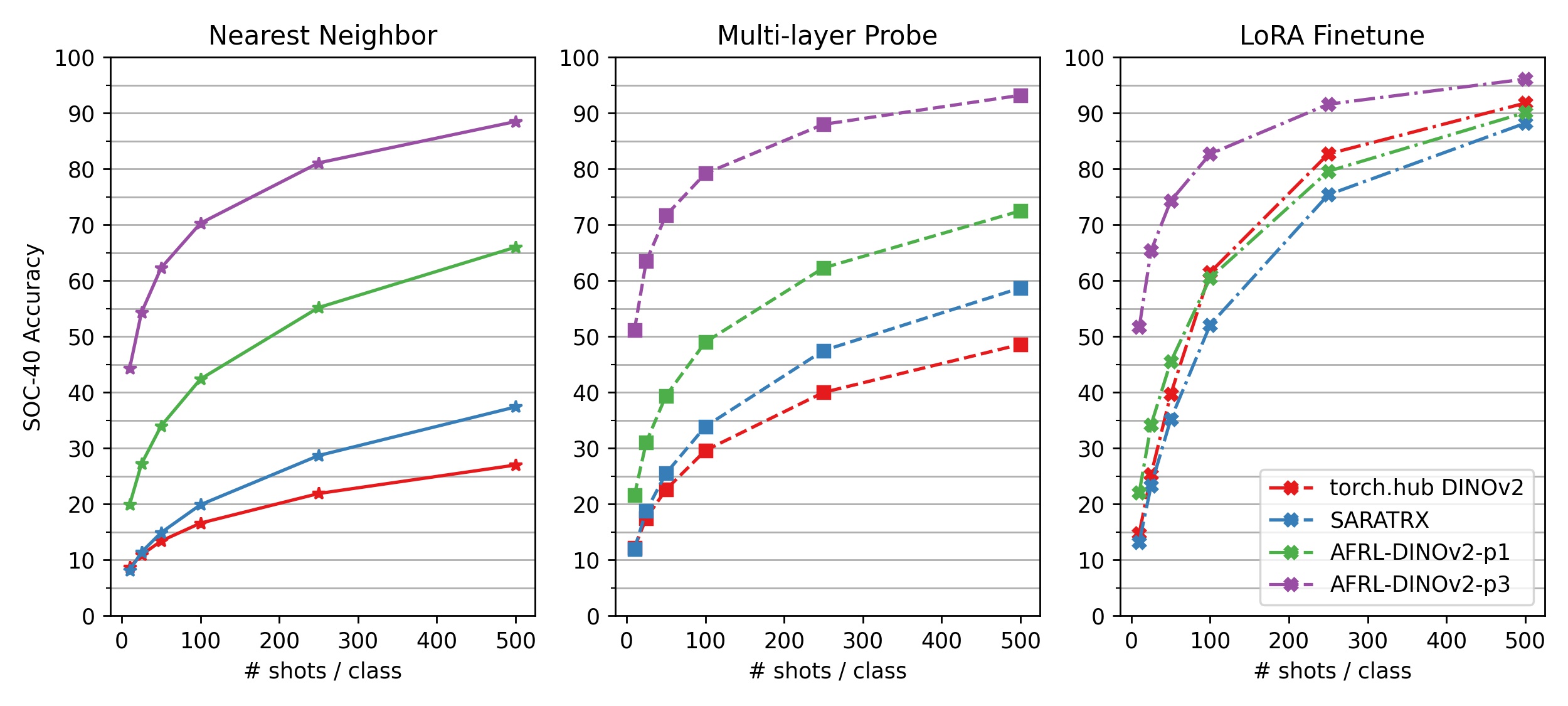}
    \caption{Few/low-shot learning results for different combinations of backbone models and task-adaptation recipes when solving a subsampled ATRNet-SOC40 task. \# shots per class for training is sampled at [10, 25, 50, 100, 250, 500]. All results are an average over 10 runs.}
    \label{fig:few-shot-soc40}
\end{figure}

A well-known benefit of using foundation models is their performance in few/low-shot data regimes.
Instead of having thousands of labeled samples per class to describe the downstream task, these models often only require tens to hundreds of labeled support samples to be performant.
The ability to work with low-labeled data in the SAR domain is particularly important because of the inherent label scarcity, especially across sensors of interest.

In this section we test the low-shot performance of foundation models on the ATRNet-SOC40 task.
We consider having [10, 25, 50, 100, 250, 500] shots per class (i.e., labeled samples per class).
For each shot value $N$ we construct and save 10 randomly sampled splits from the ATRNet-SOC40/train set and always test on the full ATRNet-SOC40/test split.
Each reported result at shot value $N$ is thus the average balanced accuracy over the \textit{same} 10 splits.
For pretrained models we use \texttt{torch.hub} DINOv2, SARATR-X, AFRL-DINOv2-p1 and AFRL-DINOv2-p3.
We also consider the effect of using different downstream task-adaptation recipes, namely Nearest Neighbor, Multi-layer Probe, and LoRA Finetuning.

Something important to keep in mind when interpreting results in this section is the assumption differences between AFRL-DINOv2-p1 and -p3. 
The -p1 backbone model has \textit{never} seen a sample from the ATRNet data distribution during SSL pretraining whereas the -p3 model has seen \textit{unlabeled} samples from this distribution during pretraining.
However, recall that -p3 is trained on the ATRNet-SOC40-train set, so these experiments still fall within an inductive learning paradigm where the test set has been completely sequestered.
With that, we can think of the -p3 model as operating in a semi-supervised regime here -- for training we have a relatively small set of labeled data and a larger set of unlabeled samples; then the test samples are completely novel.

The results are displayed in Figure~\ref{fig:few-shot-soc40} (a corresponding table with numerical results is provided in Appendix C).
The ranking of model performance with Nearest Neighbor and Multi-layer Probe heads may not be surprising as it largely echos previous experiments (AFRL-DINOv2-p3 $>$ AFRL-DINOv2-p1 $>$ SARATR-X $>$ torch.hub DINOv2).
However, what's exciting here is that the AFRL-DINOv2s show a \textit{huge} performance benefit in these low-shot tests.
With a Nearest Neighbor head at 100 shots/class, the AFRL-DINOv2-p1 outperforms SARATR-X by 22.5\% while the AFRL-DINOv2-p3 outperforms it by 50.4\%.
Similarly, with the Multi-layer Probe head the margins of -p1 and -p3 over SARATR-X at 100 shots are 15.1\% and 45.3\%, respectively.
Lastly, across all shot ranges it is generally better to use a Multi-layer Probe over a Nearest Neighbor classifier if the application requires a fixed feature extractor.

The LoRA results present a very interesting outcome which shuffle the relative model rankings we've seen before.
With a LoRA finetuning task-adaptation recipe SARATR-X is consistently the lowest performing model and the AFRL-DINOv2-p1 and \texttt{torch.hub} DINOv2 become competitive with each other.
The -p1 model has a decisive performance edge over the \texttt{torch.hub} model in the very low-shot regime (+7.3\% at 10 shots, +8.6\% at 25 shots, +5.8\% at 50 shots) while the \texttt{torch.hub} model slightly outperforms -p1 in the mid-shot regime ($\ge$100 shots).
The AFRL-DINOv2-p3 model, which is the only backbone that's seen data from the ATRNet distribution during SSL pretraining, is clearly the best performing model in these tests across the whole shot range (well over 20\% margin over the next best model at $\le$100 shots)
Overall, it's very impressive that LoRA can work in the low/few-shot regime as overfitting is usually an issue when finetuning large models on small training sets.

\section{Conclusion \& Future Work}

Motivated by the tremendous progress of visual foundation models in the wider AI/ML community our goal in this work is to update the SAR community on the applicability of this technology to the SAR domain.
We ran tests with today's most powerful frontier/foundational models, including DINOv2 \cite{oquab2024dinov}, DINOv3 \cite{simeoni2025dinov3} and PE-Core \cite{pecore} and observed their shortcomings at extracting semantically interesting features of SAR targets when used off-the-shelf.
We then showed that Self-Supervised finetuning of publicly available SSL models with SAR data is a viable path forward by training several AFRL-DINOv2s which we find to set a new state-of-the-art for SAR foundation models.
Our tests analyze the performance trade-off of using different backbones, with different downstream task-adaptation recipes, to overcome different challenges.
Overall, we highlight four important take-aways from this work that we hope will inspire/inform future efforts. 
\begin{itemize}

 \item First, DINOv2 shows to be a more powerful algorithm than the MIM-based SSL algorithm used in SARATR-X, especially for scenarios where the feature extractor is fixed downstream.
This aligns with previous observations that joint-embedding/discriminative SSL algorithms are more potent fixed feature extractors than MIM algorithms \cite{darcet2025capi, oquab2024dinov}.

 \item Second, self-supervised pretraining with SAR datasets offers numerous benefits in the SAR domain.
It allows for working with lower amounts of labeled data in the downstream tasks.
It can increase robustness to distribution shifts and extended operating conditions.
It produces models with impressive flexibility that can be used across a range of tasks involving different sensors and target sets.
And, SSL pretrained models are effective in highly restrictive deployment settings which do not provide a capability to update the feature extractor (e.g., due to hardware limitations).

 \item Third, the downstream task-adaptation recipe is very important to consider when testing SSL pretrained models.
We highly encourage future work to include recipes which keep the feature extractor fixed because it presents a relatively ``pure'' test of the expressiveness of a model's feature space.
However, we also acknowledge it's important to consider finetuning because that is where the highest top-end performance is often found.
As part of the finetuning methods, we find LoRA is a very powerful option which offers an excellent tradeoff between update complexity and performance gain, even in few/low-shot data settings.

 \item Finally, our most important takeaway from this work is that \textbf{data is gold, even if it's unlabeled}.
We are at the very beginning stages of SAR-specific foundation models, where pretraining sets are still $<$1M samples and have relatively limited target and sensor diversity.
As we repeatedly saw in this work, the AFRL-DINOv2-p3 model was the top performer because of its pretraining set. 
Without question, the fastest path to better SAR foundation models is to collect larger and more diverse SAR datasets for SSL pretraining.
   
\end{itemize}

In closing, we offer some specific suggestions for future work.
A good option for aggregating better pretraining datasets is to leverage commercial SAR data which is becoming more easily accessible online (e.g., the Umbra Open-Data program hosts a tremendous amount of free data in several formats).
Investigations into better data augmentations for SSL pretraining, including ones that appreciate SAR physics and phenomenologies, will likely lead to higher quality representations. 
Such augmentations may involve working with complex data instead of the fully processed .jpgs that are often distributed in public SAR datasets today.
We suggest that researchers be prepared to pivot as new public foundation models are released (e.g., DINOv3 and PE-Core).
Research into more advanced parameter-efficient finetuning methods may offer an even better tradeoff between update complexity and performance boost.
Finally, we encourage work in building more comprehensive benchmarks for assessing foundational-ness.

\printbibliography

\section*{Appendix}

\subsection*{A. FAQs}

\textbf{Question: Is the evaluation of AFRL-DINOv2-p3 and -p4 models on ATRNet-STAR benchmark tasks broken?}

\textit{Answer:} A careful reader may see that we have included ATRNet-SOC40/train in $\mathcal{D}_{pretrain}$ for -p3 and -p4 (in unlabeled form) and also knows from reading \cite{liu2025atrnet} that some of these data are in the test sets of the ATRNet-EOC tasks.
This creates a scenario where the backbone weights have been updated on (unlabeled) test samples from some of the downstream tasks, which is a major violation when measuring generalization performance of models trained in an \textbf{inductive learning} paradigm.
We believe it is reasonable to be on alert here.
However, we argue that this style of evaluation is permissible if we consider that unsupervised learning algorithms, including SSL, can operate in \textbf{transductive learning} settings \cite{vapnik1, vapnik2, sun19ttt}.
At a high level, inductive learning attempts to learn a general rule (i.e., model) that applies to all future samples (i.e., the test set). 
To measure the generality of the rule/model we must use a sequestered test set.
On the other hand, transductive learning attempts to adapt a model to produce the best answer for a particular set of points of interest, with less regard for creating a general rule.
Within this paradigm it is possible to execute an unsupervised learning step on a set of unlabeled test samples before rendering predictions on the set.
We believe that it is feasible to encounter both inductive and transductive learning paradigms in reality and so it is worth testing in both.
However, given how different the assumption sets are we strongly encourage that future work clearly identifies the different scenarios, e.g., using colored cells like we did in Table~\ref{tab:main-knn}.


\vskip 0.5 cm
\noindent
\textbf{Question: What about dense-prediction tasks?}

\textit{Answer:} We are aware that some foundation models are evaluated on both chip-level classification tasks and dense-prediction tasks like object detection in larger scenes.
For our purposes of assessing the applicability of existing foundation models to the SAR domain we believe that measuring performance on chip-level classification tasks is appropriate.
Intuitively, if the embedding space is not discriminative for SAR objects in classification tasks, there is no reason it will be good for detection.
We leave it to future work to assess the performance of these models on dense-prediction tasks and include the appropriate task-adaptation recipes.

\subsection*{B. More in-depth SARATR-X comparison with Liu et al.~\cite{liu2025atrnet}}

For reproducibility and consistency sake, we directly compare our ATRNet-STAR benchmark results to the ones presented in Table~5 of \cite{liu2025atrnet}.
Table~\ref{tab:compare-atrbench} shows the direct comparison. 
The top row of results are copied directly from Table~5 of \cite{liu2025atrnet}.
The latter four rows are all from our study.
For all of these results we consider a finetuning task-adaptation recipe which is what was used in \cite{liu2025atrnet}.
First, notice that our reproduction of the SARATR-X results is faithful, and actually our results are better than previously shown.
Also, notice that the AFRL-DINOv2 results are better across the board.
Not surprisingly, the -p3 models, which have been trained on unlabeled samples from the SOC-40/train set, are the best in all scenarios.
To our knowledge, these results reflect the s.o.t.a. on this benchmark in both the inductive and transductive learning paradigms.

\begin{table}[h]
\caption{Comparing our results with Table~5 of \cite{liu2025atrnet}. Recall, \colorbox{orange!20}{orange} shading in a box indicates that the feature extractor has seen \textit{unlabeled} samples from the task distribution during SSL pretraining.}
\centering
\label{tab:compare-atrbench}
\resizebox{1.\textwidth}{!}{
\begin{tabular}{lccccccc}
\toprule
Backbone                    & SOC-40 & SOC-50 & EOC-Scene & EOC-Depr & EOC-Az & EOC-Band & EOC-Pol \\ \midrule
SARATR-X (reported in \cite{liu2025atrnet}) & 96.4   & 85.2   & 19.5      & 39.9     & 26.4   & 89.2     & 84.6    \\ \midrule
SARATR-X (our reproduction) & 99.2   & 90.6   & 21.6      & 45.1     & 28.7   & 95.9     & 87.9    \\
AFRL-DINOv2-p1              & 99.2   & 91.3   & 25.2      & 46.7     & 29.1   & 97.0     & 90.5    \\
AFRL-DINOv2-p3              & \cellcolor{orange!20}99.5   & \cellcolor{orange!20}94.7   & \cellcolor{orange!20}33.9      & \cellcolor{orange!20}54.3     & \cellcolor{orange!20}43.7   & \cellcolor{orange!20}97.7     & \cellcolor{orange!20}94.7    \\
AFRL-DINOv2-p3 (LoRA)       & \cellcolor{orange!20}99.4   & \cellcolor{orange!20}94.9   & \cellcolor{orange!20}47.6      & \cellcolor{orange!20}63.8     & \cellcolor{orange!20}66.0   & \cellcolor{orange!20}97.6     & \cellcolor{orange!20}95.5   \\ \bottomrule
\end{tabular}
}
\end{table}



\subsection*{C. Numerical low-shot results}

Table~\ref{tab:few-shot} provides the numerical results used to create Figure~\ref{fig:few-shot-soc40} in the main paper.

\begin{table}[h]
\caption{Few-shot Learning Results on ATRNet SOC-40 Task}
\label{tab:few-shot}
\centering
\resizebox{0.8\textwidth}{!}{
\begin{tabular}{llccccccc}
\toprule
                                       &                   & \multicolumn{7}{c}{\# shots per class}         \\
Pretrained Ft. Extractor               & Strategy          & 10   & 25   & 50   & 100  & 250  & 500  & ALL  \\ \midrule
\multirow{3}{*}{torch.hub DINOv2-ViTB} & Nearest Neighbor          & 8.7  & 10.9 & 13.4 & 16.6 & 21.9 & 27   & 38.7 \\
                                       & Multi-layer Probe & 12.2 & 17.5 & 22.6 & 29.6 & 40   & 48.6 & 63.7 \\
                                       & LoRA Finetuning        & 14.8 & 25.3 & 39.7 & 61.5 & 82.7 & 91.8 & 98.8 \\ \midrule
\multirow{3}{*}{SARATRX-notest}        & Nearest Neighbor          & 8.1  & 11.4 & 14.9 & 19.9 & 28.7 & 37.4 & 56.6 \\
                                       & Multi-layer Probe & 12   & 18.8 & 25.5 & 33.9 & 47.5 & 58.7 & 74.8 \\
                                       & LoRA Finetuning        & 13.2 & 23.3 & 35.2 & 52.1 & 75.4 & 88.2 & 98.1 \\ \midrule
\multirow{3}{*}{AFRL-DINOv2-ViTB-p1}   & Nearest Neighbor          & 19.9 & 27.2 & 34   & 42.4 & 55.2 & 66   & 84.1 \\
                                       & Multi-layer Probe & 21.6 & 31.1 & 39.4 & 49   & 62.3 & 72.5 & 87.1 \\
                                       & LoRA Finetuning       & 22.1 & 34.2 & 45.5 & 60.5 & 79.6 & 90.1 & 98.6 \\ \midrule
\multirow{3}{*}{AFRL-DINOv2-ViTB-p3}   & Nearest Neighbor          & 44.3 & 54.3 & 62.3 & 70.3 & 81.1 & 88.5 & 96.7 \\
                                       & Multi-layer Probe & 51.2 & 63.5 & 71.7 & 79.2 & 88   & 93.2 & 98.3 \\
                                       & LoRA Finetuning        & 51.7 & 65.4 & 74.3 & 82.7 & 91.6 & 96.1 & 99.4 \\ \bottomrule
\end{tabular}
}
\end{table}

\end{document}